\documentclass[journal]{IEEEtran}

\usepackage{cite}
\usepackage{amsmath,amssymb,amsfonts}
\usepackage{graphicx}
\usepackage{algorithm}        
\usepackage{algorithmic}      
\usepackage[caption=false,font=footnotesize]{subfig} 
\usepackage{xcolor}
\usepackage{float}
\usepackage{textcomp}
\usepackage{url} 
\usepackage[hidelinks]{hyperref}
\usepackage{stfloats}         
\usepackage{balance}          
\usepackage[utf8]{inputenc} 
\usepackage{newunicodechar}
\newunicodechar{≈}{\approx}
\usepackage[T1]{fontenc}
\usepackage{textcomp}
\usepackage{newtxtext,newtxmath} 

\usepackage{tabularx,array}

\begin{document}

\title{Edge-AI Perception Node for Cooperative Road-Safety Enforcement and Connected-Vehicle Integration}
\author{Shree~Charran~R and Rahul~Kumar~Dubey,~\textit{Senior~Member,~IEEE}%
\thanks{Shree~Charran~R is with the Indian Institute of Science (IISc), Bengaluru~560012, India (e-mail: \href{mailto:shreer@alum.iisc.ac.in}{shreer@alum.iisc.ac.in}).}%
\thanks{Rahul~Kumar~Dubey is with \textit{Bosch Global Software Technologies Private Limited (BGSW)}, Bengaluru~560095, India (e-mail: \href{mailto:RahulKumar.Dubey@in.bosch.com}{RahulKumar.Dubey@in.bosch.com}).}%
}

\maketitle

\begin{abstract}
Rapid motorization in emerging economies such as India has created severe enforcement asymmetries—over \textbf{11~million recorded violations in~2023} against a human policing density of roughly one officer per 4{,}000 vehicles. Traditional surveillance and manual ticketing cannot scale to this magnitude, motivating the need for an \textbf{autonomous, cooperative, and energy-efficient edge-AI perception infrastructure}. This paper presents a \textbf{real-time roadside perception node} for multi-class traffic-violation analytics and safety-event dissemination within a \textit{connected and intelligent-vehicle ecosystem}. The node integrates \textbf{YOLOv8-Nano} for high-accuracy multi-object detection, \textbf{DeepSORT} for temporally consistent vehicle tracking, and a \textbf{rule-guided OCR–post-processing engine} capable of recognizing degraded or multilingual license plates compliant with MoRTH~AIS-159 and ISO~7591 visual-contrast standards. Deployed on an NVIDIA~Jetson~Nano (128-core Maxwell GPU) and optimized via TensorRT~FP16 quantization, the system sustains \textbf{28–30~fps inference at~9.6~W}, achieving \textbf{97.7\% violation-detection accuracy} and \textbf{84.9\% OCR precision} across five violation classes—signal-jumping, zebra-crossing breach, wrong-way driving, illegal U-turn, and speeding—without manual ROI calibration. Comparative benchmarking against YOLOv4-Tiny, PP-YOLOE-S, and NanoDet-Plus demonstrates a~\textbf{10.7\%~mAP gain} and~\textbf{1.4×~accuracy-per-watt improvement}. Beyond enforcement, the node publishes standardized safety events (\textit{CAM/DENM-type}) to connected vehicles and ITS back-ends via V2X protocols, demonstrating that \textbf{roadside edge-AI analytics can augment cooperative perception and proactive road-safety management} within the IEEE~Intelligent~Vehicles~ecosystem.
\end{abstract}

\begin{IEEEkeywords}
Edge~AI, Cooperative~Perception, Intelligent~Vehicles, V2X~Communication, YOLOv8, DeepSORT, OCR, TensorRT, Road~Safety
\end{IEEEkeywords}

\section*{Nomenclature}

\begin{IEEEdescription}[\IEEEusemathlabelsep\IEEEsetlabelwidth{$\mathcal{Z}_{\text{violation}}$}]
\item[$F_t$] Input video frame at time~$t$.
\item[$\mathcal{D}_t$] Set of detected objects in frame~$t$ from YOLOv8.
\item[$\mathcal{R}$] Region-of-interest (ROI) masks for violation zones automatically derived from static geometry.
\item[$N$] Number of frames between virtual ``start'' and ``stop'' line crossings.
\item[$T_F$] Frame interval or time per frame (s).
\item[$d$] Distance between the two virtual speed lines (m).
\item[$v$] Estimated vehicle speed (km/h).
\item[$\mathcal{H}$] Convex hull computed over static scene geometry (lanes, dividers, zebra crossings).
\item[$\mathcal{M}_{\text{static}}$] Set of static road features detected in the scene.
\item[$\mathcal{Z}_{\text{violation}}$] Geometric violation zones (e.g., stop line, divider region) derived from $\mathcal{M}_{\text{static}}$.
\item[$\text{mAP}@0.5$] Mean average precision at 0.5~IoU threshold.
\item[$\text{FPS}$] Frames per second — inference throughput metric.
\item[$P_{\text{GPU}}$] Power consumption of the Jetson~Nano GPU (W).
\item[$\text{OCR}_{\text{acc}}$] License-plate recognition accuracy of the OCR module.
\item[$\text{IoU}$] Intersection-over-Union — overlap ratio between prediction and ground truth.
\item[$\text{FP16}$] Half-precision floating-point inference mode used in TensorRT optimization.
\end{IEEEdescription}

\IEEEpeerreviewmaketitle

\section{Introduction}
\IEEEPARstart{R}{oad} safety enforcement and cooperative perception remain persistent challenges in emerging economies, particularly in dense and unstructured urban centers such as Mumbai, Delhi, and Bengaluru. The World Health Organization reports that India accounts for nearly \textbf{11\% of global road fatalities}, exceeding 150{,}000 deaths annually, with violation-induced accidents contributing to almost 70\% of cases. In 2023, Indian traffic authorities recorded over \textbf{11~million documented violations}, an increase of nearly 30\% year-on-year~\cite{R1,R2,R3}. Despite stricter penalties under the 2019 Motor Vehicles Amendment Act and the introduction of electronic challan systems, enforcement capacity remains severely constrained: more than \textbf{330~million registered vehicles} but only \textbf{80{,}000 active traffic officers}~\cite{R4,R5}, corresponding to roughly one officer per 4{,}000 vehicles. This imbalance highlights the need for scalable, automated, and cooperative safety analytics capable of ensuring compliance without continuous human oversight.

Existing camera-based enforcement and surveillance platforms—often proprietary and cloud-dependent—are tailored to structured, high-income urban layouts and rely on fixed geometric configurations. Their deployment in Indian cities is challenged by \textbf{non-standardized intersections, heterogeneous traffic composition (two-wheelers, autos, pedestrians)}, and intermittent connectivity. Furthermore, rule-based or static-region video pipelines fail to generalize across lighting and topology variations, producing high false-positive rates and costly recalibration cycles.

Recent advances in \textbf{edge artificial intelligence (Edge-AI)} and embedded computer vision enable real-time perception directly at the roadside. Compact systems such as NVIDIA~Jetson~Nano/Orin, Google~Coral~TPU, and Intel~Movidius perform deep inference locally within a 5–10~W power envelope, thereby minimizing latency, preserving privacy, and reducing network dependence. When connected via \textbf{V2X or ITS-G5 links}, these edge nodes can act as \textbf{cooperative perception agents}—sharing structured safety events (e.g., red-light violations, overspeed warnings) with nearby vehicles or traffic controllers to facilitate proactive driver assistance and adaptive signal control. Edge computing also aligns with global sustainability goals (UN~SDG~11.2 and~3.6) by supporting decentralized, low-energy mobility infrastructure.

A preliminary version of this research focusing on two-wheeler violations using a YOLOv4 + DeepSORT + Tesseract pipeline was reported in~\cite{R10}. That work demonstrated feasibility under controlled lighting and camera geometries. The present study substantially extends that foundation by introducing an anchor-free \textbf{YOLOv8} architecture, automatic ROI generation, and TensorRT-accelerated deployment on an NVIDIA~Jetson~Nano. In addition to expanding coverage to multiple vehicle types and violation classes, the system now generates standardized \textbf{V2X safety messages (CAM/DENM-type)} for integration with cooperative ITS dashboards and connected-vehicle ecosystems. These advancements transform the earlier CPU-bound prototype into a fully edge-resident, regulation-aligned perception node for heterogeneous Indian traffic conditions.

\textbf{Motivated by these gaps, this paper proposes a real-time, low-cost, and cooperative edge-AI perception framework} for multi-violation detection, analytics, and event dissemination in connected-vehicle environments. Unlike monolithic cloud systems or single-violation detectors, the proposed node unifies modern deep detectors, multilingual OCR, and automatic region-of-interest (ROI) generation into a hardware-efficient pipeline deployable even at low-cost city intersections. By leveraging TensorRT optimizations and on-device post-processing, the framework ensures real-time inference, energy efficiency, and privacy preservation while maintaining compatibility with ITS back-end APIs and V2X communication standards.

The major contributions of this research are summarized below:
\begin{itemize}
    \item \textbf{Anchor-Free Detection Pipeline:} Integration of the \textbf{YOLOv8-Nano} architecture for high-accuracy, real-time object detection on embedded GPUs, replacing legacy YOLOv4 pipelines and achieving substantial accuracy–latency gains.
    \item \textbf{Rule-Guided Multilingual OCR:} Development of a lightweight OCR engine compliant with Indian license-plate standards (MoRTH~AIS-159) and multilingual scripts, ensuring robust recognition under motion blur and variable illumination.
    \item \textbf{Unified Multi-Violation Analytics:} End-to-end detection of five common violations—signal jumping, zebra-crossing breach, wrong-way driving, illegal U-turn, and speeding—\textit{without manual ROI calibration}, enhancing scalability and interoperability.
    \item \textbf{Edge Deployment with Cooperative Output:} Demonstration of real-time operation (28–30~fps at 9.6~W) on the \textbf{NVIDIA~Jetson~Nano}, publishing structured V2X safety events to ITS controllers, confirming feasibility for city-scale cooperative-perception deployments.
\end{itemize}

The remainder of this paper is organized as follows. Section~\ref{sec:background} reviews prior work on vehicle perception, OCR-based license-plate recognition, and cooperative edge-AI systems. Section~\ref{sec:methodology} describes the proposed architecture and algorithms. Section~\ref{sec:experimental_setup} details the datasets, experimental design, and hardware configuration. Section~\ref{sec:results} presents quantitative evaluation and comparative analysis. Finally, Section~\ref{sec:Conclusion} concludes with key findings, industrial implications, and directions for future research.
\section{Background}
\label{sec:background}

\subsection{Vehicle Perception and Violation Detection}

Early traffic-safety and enforcement systems relied on handcrafted image-processing techniques and first-generation deep detectors. Arnob \textit{et al.}~\cite{R1} combined Canny edge detection with the Hough Transform to identify lane-change violations, followed by a YOLO-based plate extraction stage. Although effective under controlled conditions, this multi-stage pipeline incurred high computational cost and lacked scalability for real-time city deployment.  

Taheri \textit{et al.}~\cite{R2} introduced an edge-optimized Tiny-YOLOv3 variant by pruning convolutional layers and removing batch normalization, achieving a 12.7 k-parameter reduction and improved throughput. Yet inference on the NVIDIA Jetson Nano plateaued at 17 fps—insufficient for dense traffic where each frame may contain 30 + objects.  

Subsequent work improved robustness under variable illumination and occlusion. Charran \textit{et al.}~\cite{R10} implemented a YOLOv4–DeepSORT pipeline for two-wheeler violations such as helmet non-compliance and phone use, integrating cropped plate recognition for ticket generation. Franklin \textit{et al.}~\cite{R11} used YOLOv3 for red-light enforcement but relied on manually defined regions of interest (ROIs), limiting cross-location portability. Rathore \textit{et al.}~\cite{R12} deployed a GPU-based SSD on an in-vehicle fog node to detect U-turn and divider violations; performance degraded under glare and camera-angle variations.  

Table~\ref{tab:comparison_existing_yesno} summarizes representative approaches for automated violation detection and license-plate recognition on embedded or edge platforms.  
Recent detector families—\textbf{YOLOv8} (2023), \textbf{YOLOv11} (2024), and \textbf{PP-YOLOE} (2022)—achieve superior accuracy–latency trade-offs via the \textbf{C2f backbone} for efficient gradient flow, \textbf{SPPF} for multi-scale context, and \textbf{decoupled anchor-free heads}.  
Their \textit{nano} and \textit{small} variants are optimized for embedded-GPU inference, outperforming legacy YOLOv3/v4 models in both mAP and energy efficiency—making them ideal for \textbf{edge perception nodes} deployed within cooperative ITS or V2X frameworks.  
Compact detectors including YOLOv4-Tiny~\cite{R15}, YOLOv8n~\cite{R18}, PP-YOLOE-S~\cite{R19}, YOLOv11n~\cite{R20}, and NanoDet-Plus~\cite{R21} were benchmarked on the Jetson Nano (Table~\ref{tab:detector_comparison}), confirming feasibility for 10 W, near-real-time sensing.

\begin{table*}[t]
\centering
\caption{Comparative Assessment of Existing Traffic-Violation Detection Approaches}
\label{tab:comparison_existing_yesno}
\renewcommand{\arraystretch}{1}
\small
\begin{tabularx}{\textwidth}{|p{1.1cm}|*{7}{>{\centering\arraybackslash}X|}}
\hline
\textbf{Reference} & \textbf{Edge Deployment} & \textbf{Multi-Violation} & \textbf{Auto ROI} & \textbf{OCR Integration} & \textbf{Real-Time ($\geq$ 30 fps)} & \textbf{Low Power ($<10$ W)} & \textbf{Accuracy ($\geq$ 90 \%)} \\ \hline
\cite{R1}  & No      & No            & No & No  & No               & No        & No      \\ \hline
\cite{R2}  & Yes     & No            & No & No  & Partial (17 fps)  & Yes       & No      \\ \hline
\cite{R7}–\cite{R9} & Yes & No & No & Yes & No & Yes & No \\ \hline
\cite{R10} & Partial & Yes (2-W only) & No & Yes & Partial (20 fps) & No & $\sim$85 \% \\ \hline
\cite{R11} & No & No & No & No & No & No & No \\ \hline
\cite{R12} & Yes & Yes (2 types) & No & No & No & No & $\sim$80 \% \\ \hline
\textbf{Proposed} & \textbf{Yes} & \textbf{Yes (5 types)} & \textbf{Yes} & \textbf{Yes} & \textbf{Yes (30 fps)} & \textbf{Yes (9.6 W)} & \textbf{97.7 \%} \\ \hline
\end{tabularx}
\end{table*}

\begin{table}[H]
\centering
\caption{Representative object detectors for edge deployment (Jetson Nano, 640×640 input).}
\label{tab:detector_comparison}
\begin{tabular}{lccp{3.2cm}}
\hline
\textbf{Model} & \textbf{mAP@0.5} & \textbf{FPS} & \textbf{Comments} \\ \hline
YOLOv4-Tiny (2020) & 38.2 & 21 & Early edge baseline. \\
PP-YOLOE-S (2022) & 46.7 & 24 & Balanced accuracy–speed. \\
YOLOv8n (2023) & 48.9 & 28 & C2f + SPPF; TensorRT-optimized. \\
YOLOv11n (2024) & 51.3 & 30 + & Transformer-neck; SoA efficiency. \\
NanoDet-Plus (2022) & 43.5 & 32 & Ultra-lightweight; lower precision. \\ \hline
\end{tabular}
\end{table}

These trends show that sub-50 MB detectors can now sustain real-time throughput on $\approx 10$ W hardware, paving the way for scalable \textbf{roadside cooperative-perception nodes} supporting city-level connected-safety services.

\subsection{License-Plate Recognition on Edge Devices}

Embedded license-plate recognition (LPR) traditionally follows two stages—region segmentation and optical character recognition (OCR). Firasanti \textit{et al.}~\cite{R7} compared Canny and Otsu thresholding on a Raspberry Pi, observing that Canny achieved higher accuracy but required illumination > 500 lux. Abirami \textit{et al.}~\cite{R8} and Kumthekar \textit{et al.}~\cite{R9} implemented OpenCV-based segmentation with Tesseract OCR but were highly sensitive to vibration, font variation, and low-resolution CCTV feeds.  

From an Indian-deployment perspective, classical LPR pipelines face non-standard plate fonts, partial occlusion, and multilingual scripts. Few works enforce \textbf{MoRTH AIS-159} or \textbf{ISO 7591} visibility standards, reducing robustness under poor lighting. Moreover, generic OCR engines rarely perform syntactic validation of vehicle numbers (\texttt{AA00AA0000}), causing high false recognition under motion blur.  

Edge-oriented LPR systems must therefore balance three competing objectives: inference speed, recognition accuracy, and model compactness. Modern lightweight OCR frameworks (EasyOCR, PaddleOCR, Tesseract 5) offer multilingual support and INT8/FP16 inference, yet consistent field reliability on embedded GPUs remains challenging—especially for cooperative nodes that must deliver standardized plate data to V2X networks in real time.

\subsection{Identified Gaps and Cooperative-Perception Motivation}

A review of prior literature highlights several gaps motivating this study:
\begin{itemize}
    \item \textbf{Isolated-Task Focus:} Most works address a single violation type or standalone OCR without an integrated perception-to-action chain useful for connected vehicles.
    \item \textbf{Legacy Detection Pipelines:} YOLOv3/v4 models still dominate despite anchor-free and transformer-based variants that offer higher accuracy at lower latency.
    \item \textbf{Manual ROI Configuration:} Hand-drawn ROIs remain common, limiting scalability across intersections and camera geometries.
    \item \textbf{Lack of Cooperative Outputs:} Few systems publish structured safety messages (e.g., CAM/DENM) that can feed adaptive signal controllers or in-vehicle ADAS modules.
    \item \textbf{Incomplete Edge Integration:} Very few demonstrate a fully integrated detection–tracking–OCR stack on compact, low-power hardware suitable for heterogeneous, real-world conditions.
\end{itemize}

Addressing these gaps, the present work introduces a unified, real-time, multi-violation perception node that combines \textbf{YOLOv8} for visual sensing, \textbf{DeepSORT} for spatiotemporal tracking, and a \textbf{rule-enhanced OCR} pipeline optimized for the \textbf{Jetson Nano}.  
The node eliminates manual ROI calibration, operates at 30 fps under 10 W, and exports violation events as standardized cooperative-safety packets, forming a \textit{scalable, regulation-compliant edge-AI perception element for connected and intelligent vehicles}.

\section{Proposed Methodology}
\label{sec:methodology}

\begin{figure*}[t]
\centering
\includegraphics[width=\linewidth]{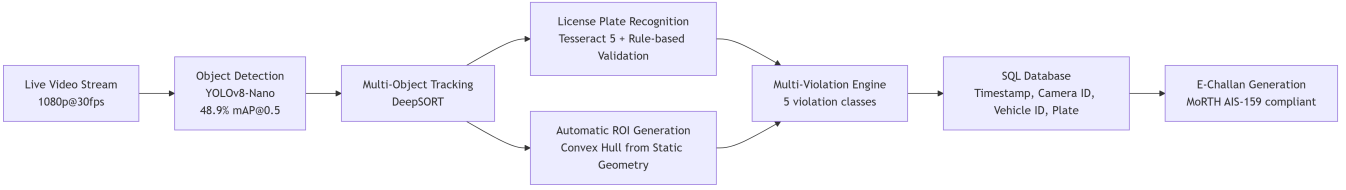}
\caption{System architecture of the roadside \emph{cooperative perception node}: on-device detection, tracking, OCR, violation reasoning, and standardized safety-event publishing to V2X/ITS back-ends.}
\label{fig:system_architecture}
\end{figure*}

This section presents the proposed end-to-end \textbf{edge-resident roadside perception node} for real-time, automated multi-violation analytics \emph{with cooperative outputs to connected vehicles and ITS back-ends}. The modular architecture in Fig.~\ref{fig:system_architecture} targets low-power embedded GPUs and comprises five subsystems:
(i)~object detection, (ii)~multi-object tracking, (iii)~license-plate detection and recognition, (iv)~violation reasoning and logging, and (v)~\textbf{cooperative output \& V2X interface}. Together, these components form a scalable, privacy-preserving perception element for \emph{cooperative road-safety} in heterogeneous urban environments.

\begin{itemize}
  \item \textbf{Object Detection:} Each incoming frame from a live stream is processed by a YOLOv8-Nano detector to identify vehicles, zebra crossings, lanes, and license plates.
  \item \textbf{Object Tracking:} Detected vehicles are assigned persistent IDs by DeepSORT, enabling spatiotemporal consistency and preventing duplicate counts.
  \item \textbf{License-Plate Recognition:} Cropped plate regions are decoded via OCR with rule-based syntax validation for robust alphanumeric recognition.
  \item \textbf{Violation Reasoning \& Logging:} Confirmed events are materialized in an SQL store (timestamp, track ID, class, plate hash), supporting analytics and audit.
  \item \textbf{Cooperative Output \& V2X:} Safety events are published as compact messages (CAM/DENM-style payloads) over MQTT/ITS-G5/C-V2X gateways to \emph{connected vehicles and ITS controllers}.
\end{itemize}

All algorithms are optimized for real-time operation on resource-constrained platforms such as the \textbf{NVIDIA Jetson Nano (4~GB, 128-core Maxwell GPU)}. The unified pipeline eliminates cloud dependence, preserves privacy, and maintains sub-50~ms end-to-end latency—critical for \emph{cooperative} safety applications as well as enforcement.

\subsection{Object Detection — YOLOv8}
The detection module employs YOLOv8-Nano~\cite{R15}, balancing accuracy and compute for embedded hardware. YOLOv8 uses an \textbf{anchor-free, decoupled head} and \textbf{C2f backbone} (CSP-inspired) to improve gradient flow and feature reuse. The model is fine-tuned on COCO + a custom Indian-traffic set (80/20 split) using SGD (lr~0.01, batch~16, 150~epochs). In TensorRT FP16 mode it achieves 28–30~fps at 48.9\% mAP@0.5 under a $<10$~W envelope.

\begin{itemize}
  \item \textbf{Backbone:} C2f layers with Spatial Pyramid Pooling–Fast (SPPF) for multi-scale context.
  \item \textbf{Neck:} Path-Aggregation Feature Pyramid (PAFPN) for scale-robust features.
  \item \textbf{Head:} Anchor-free, decoupled branches for classification and box regression producing $(x,y,h,w)$ and confidences.
\end{itemize}

Quantization-aware training and TensorRT graph fusion shrink the detector footprint (e.g., 36~MB vs. 61~MB in YOLOv4-Tiny) and raise throughput within the Nano’s thermal budget. Although YOLOv11n~\cite{R20} offers modest mAP gains (+2–3\%), profiling on Jetson Nano showed a 1.4$\times$ larger model ($\approx$56~MB) and 15–20\% higher power draw, stressing the 10~W target and with partial export support on JetPack~4.6. Hence, \textbf{YOLOv8-Nano} is selected for \emph{reproducible, energy-efficient deployment}.

\subsection{Automatic ROI Generation}
Manual regions of interest (ROIs) impede scale across intersections. We derive ROIs automatically from static elements (lanes, dividers, zebra crossings) detected by YOLOv8, enabling \emph{plug-and-play} deployment.

\begin{algorithm}[!t]
\caption{Automatic ROI Generation for Violation Zones}
\label{alg:auto_roi}
\begin{algorithmic}[1]
\REQUIRE Frame sequence $\{F_t\}$; detections $\{\mathcal{D}_t\}$
\ENSURE ROI masks $\mathcal{R}$ for zebra crossings, lanes, dividers
\FOR{each $F_t$}
  \STATE Select static classes: \texttt{zebra\_crossing}, \texttt{lane}, \texttt{divider}
  \STATE Compute convex-hull polygons enclosing selected detections
  \STATE Temporal-average hull vertices over $K$ frames to reduce jitter
  \STATE Rasterize binary masks $\mathcal{R}$ aligned to averaged polygons
\ENDFOR
\STATE Persist $\mathcal{R}$ for geometric/temporal checks
\end{algorithmic}
\end{algorithm}

For \emph{signal-jump}, the upper edge of the zebra polygon defines a virtual \textbf{stop line}; a violation triggers when a tracked vehicle crosses during red.  
For \emph{wrong-way}, estimate the nominal lane vector $\vec{v}_n$ from lane centroids; any vehicle motion $\vec{v}_i$ with $\vec{v}_i\!\cdot\!\vec{v}_n\!<\!0$ is flagged.  
For \emph{speeding}, two virtual lines (\emph{Start}, \emph{Stop}) spaced by distance $d$ yield
\begin{equation}
    v \;=\; \frac{d}{N \, T_F},
\end{equation}
where $N$ is the frame count between crossings and $T_F$ the frame interval (s).  
This adaptive ROI mechanism ensures geometric consistency and rapid commissioning in unstructured scenes.

\begin{figure}[t]
\centering
\includegraphics[width=\columnwidth]{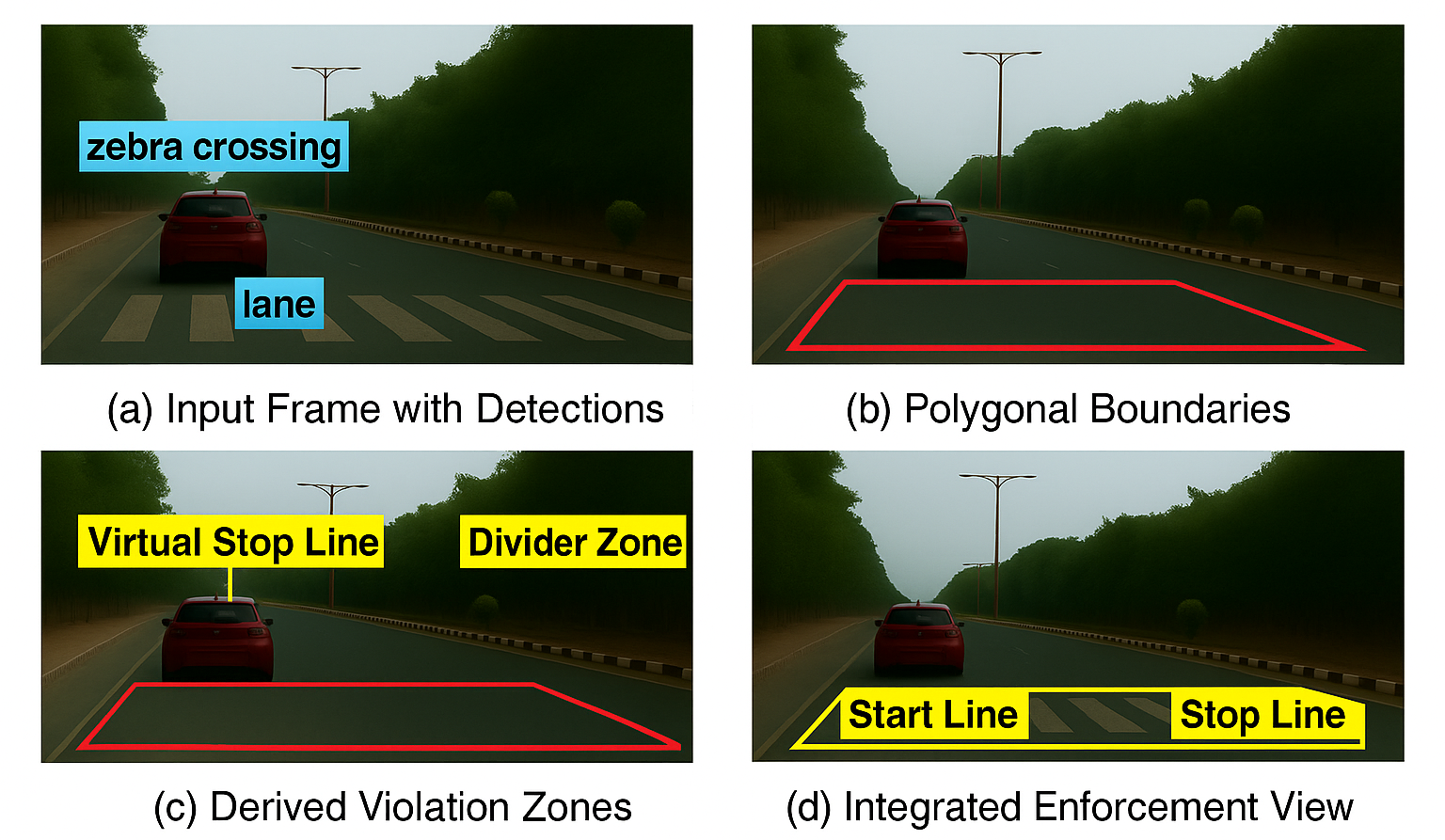}
\caption{Automatic ROI generation: (a) static detections; (b) convex-hull boundaries; (c) derived violation zones (virtual stop/divider lines); (d) integrated visualization.}
\label{fig:roi_generation}
\end{figure}

\subsection{Object Tracking — DeepSORT}
DeepSORT~\cite{R16} couples a Kalman filter with CNN appearance embeddings for robust re-identification. Each object receives a persistent ID through short-term occlusions, enabling:
\begin{itemize}
  \item \textbf{Temporal continuity} across frames,
  \item \textbf{Occlusion recovery} and re-association,
  \item \textbf{Cross-event aggregation} into vehicle-level records.
\end{itemize}
This yields $>\!98\%$ event-aggregation accuracy in moderate congestion and reduces duplicate alarms.

\subsection{License-Plate Detection and Recognition — YOLOv8 + OCR}
The LPR pipeline combines YOLOv8 localization with lightweight OCR:
\begin{itemize}
  \item \textbf{Detection:} A YOLOv8 sub-model trained on 11{,}271 Indian plates localizes regions for cropping.
  \item \textbf{Preprocessing:} Grayscale, bilateral filtering, adaptive thresholding, and CLAHE to stabilize contrast.
  \item \textbf{Recognition \& Validation:} Tesseract~5~\cite{R17} with regex-based syntactic checks (\texttt{AA00AA0000}); multi-frame confidence voting over $T$ frames mitigates transient errors, achieving 84.9\% mean accuracy under mixed illumination.
\end{itemize}
The OCR stack ($<\!15$~MB) runs on CPU cores, overlapping with GPU inference.

\subsection{Violation Reasoning}
Four principal violation classes are inferred by geometric and temporal logic:
\begin{enumerate}
  \item \textbf{Illegal U-turn:} Divider openings form three zones (A,B,C) at $\pm 90^\circ$; rapid traversal of all zones implies a U-turn event.
  \item \textbf{Signal Jump / Zebra Breach:} Virtual stop-line tests with signal phase or synchronized timing.
  \item \textbf{Wrong-Way Driving:} $\vec{v}_i\!\cdot\!\vec{v}_n\!<\!0$ indicates inverse flow.
  \item \textbf{Speeding:} Start/stop line timing via Eq.~(1) with calibrated spacing $d$.
\end{enumerate}

\begin{figure*}[t]
  \centering
  \subfloat[Illegal U-turn]{%
    \includegraphics[width=0.48\linewidth,height=0.28\linewidth,keepaspectratio]{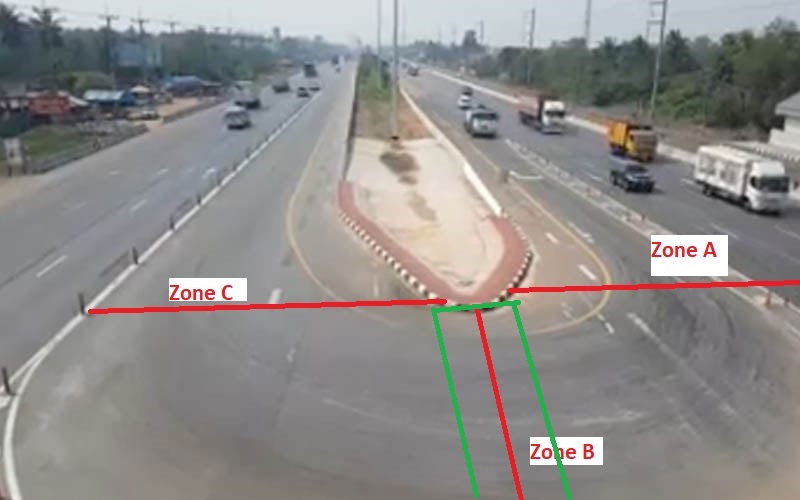}}
  \hfill
  \subfloat[Signal jump]{%
    \includegraphics[width=0.48\linewidth,height=0.28\linewidth,keepaspectratio]{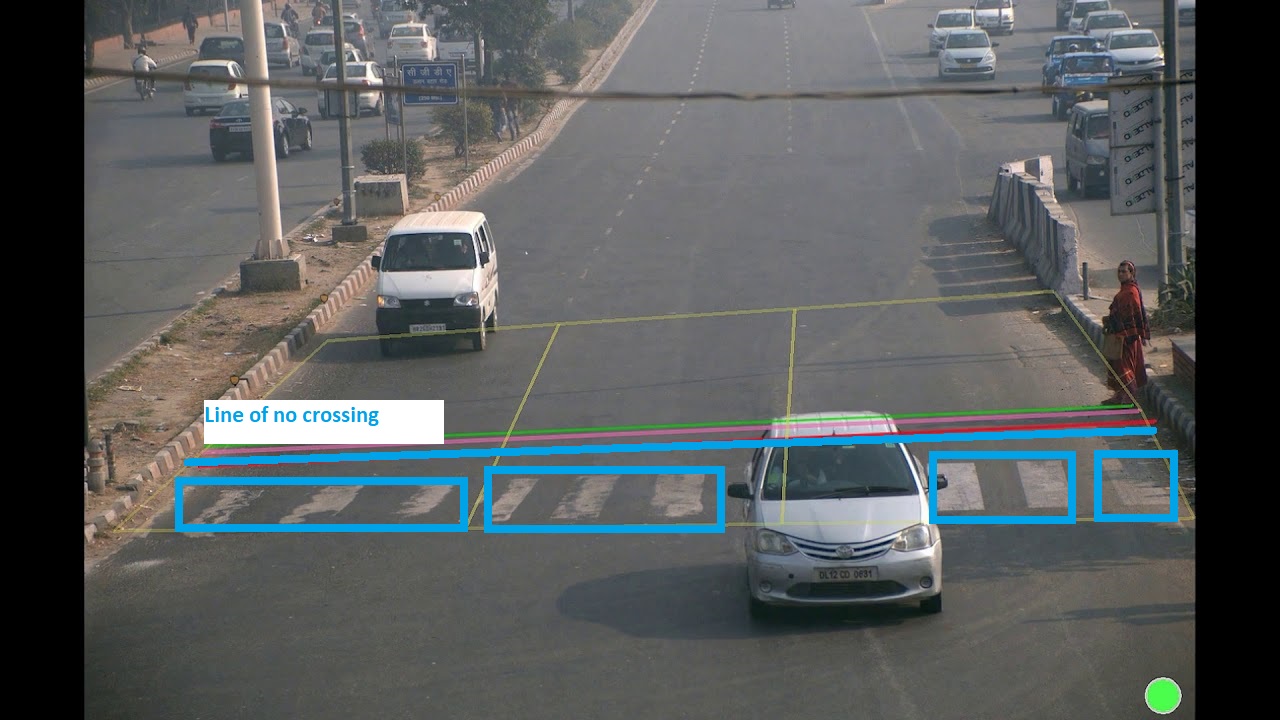}}
  \vspace{4pt}
  \subfloat[Wrong-way]{%
    \includegraphics[width=0.48\linewidth,height=0.28\linewidth,keepaspectratio]{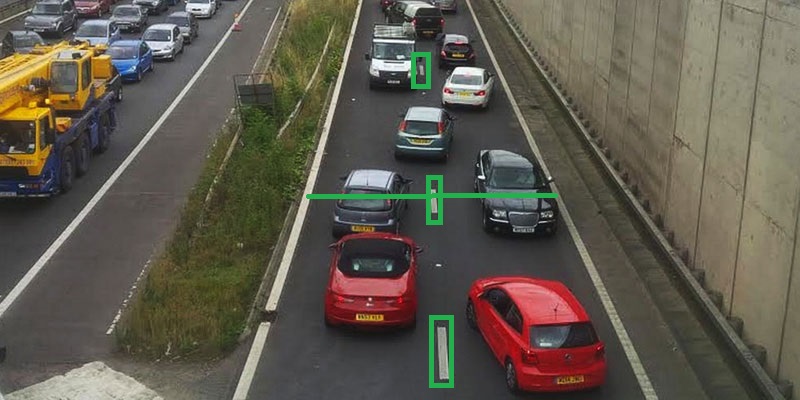}}
  \hfill
  \subfloat[Speeding]{%
    \includegraphics[width=0.48\linewidth,height=0.28\linewidth,keepaspectratio]{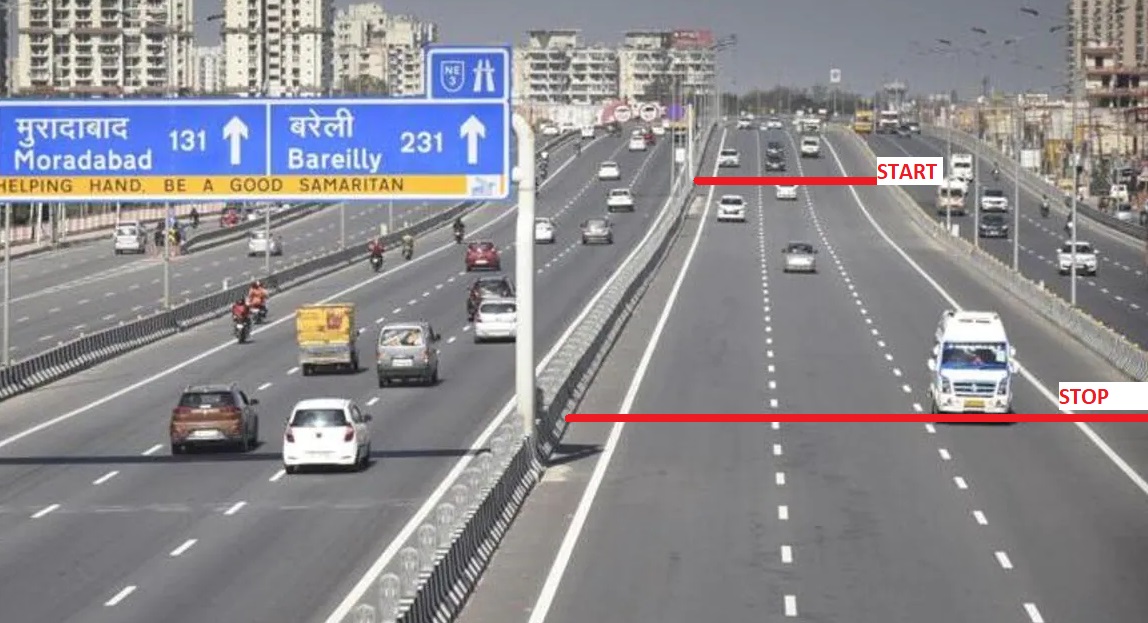}}
  \caption{Examples of detected violations: (a) Illegal U-turn, (b) Signal jump, (c) Wrong-way driving, and (d) Speeding.}
  \label{fig:violations_examples}
\end{figure*}

Each verified event is recorded with timestamp, camera ID, track ID, plate hash, and violation class. These logs support \emph{spatiotemporal heat maps}, \emph{peak-hour profiling}, and \emph{policy analytics}.

\subsection{Cooperative Output and V2X/ITS Interface}
\label{sec:v2x_interface}
To enable \emph{cooperative safety}, confirmed events are exported as compact messages compatible with CAM/DENM-style semantics:

\noindent\textbf{Payload (binary/JSON):}
\begin{itemize}
  \item \texttt{msg\_type} (e.g., \texttt{VIOL\_RL}, \texttt{VIOL\_SPD}, \texttt{VIOL\_WW}), \texttt{severity}, \texttt{confidence}
  \item \texttt{t\_utc}, \texttt{lat}, \texttt{lon}, \texttt{heading}, \texttt{speed}
  \item \texttt{track\_id} (ephemeral), \texttt{plate\_hash} (salted SHA-256), \texttt{cam\_id}
  \item \texttt{roi\_id}, \texttt{evidence\_uri} (local/edge URL) \emph{(optional)}
\end{itemize}

\noindent\textbf{Transport and timing:}
\begin{itemize}
  \item \textbf{Transport:} MQTT over TLS~1.2 to an ITS message broker; alternatively, an ITS-G5/C-V2X gateway relays CAN/DENM-like PDUs to OBUs/RSUs.
  \item \textbf{Rate-limiting:} Max 10~Hz per node with deduplication windows (e.g., 3–5~s) to prevent alert storms.
  \item \textbf{Time sync:} NTP/PTP for sub-10~ms stamping; GPS time as fallback for isolated nodes.
\end{itemize}

\noindent\textbf{Privacy \& security:}
\begin{itemize}
  \item \textbf{Minimization:} Only \emph{hashed} identifiers leave the node; raw plates remain on-device unless escrow is legally requested.
  \item \textbf{TLS + mTLS:} Node and broker use mutual authentication; keys rotate periodically.
  \item \textbf{Data retention:} Rolling buffer (e.g., 7–30 days) with encrypted storage; WGS-84 geotagging for forensics.
\end{itemize}

This interface allows the perception node to act as a \emph{cooperative safety sensor}: vehicles receive timely warnings (e.g., red-light running hotspot ahead), and traffic controllers incorporate violation analytics into adaptive strategies—closing the loop between \emph{perception, V2X dissemination, and ITS actuation}.

\section{Experimental Setup}
\label{sec:experimental_setup}

\subsection{Hardware}
The node was deployed on an \textbf{NVIDIA Jetson Nano Developer Kit}—a cost-effective embedded-GPU platform widely used for edge perception in intelligent-vehicle and ITS prototypes. The device integrates a 128-core Maxwell GPU (CUDA/cuDNN) with a quad-core ARM~A57 CPU in a compact, fan-cooled module. Its low thermal design power (\textbf{10~W}) and small form factor (100$\times$80~mm) make it suitable for pole-mounted roadside units (RSUs) operating from solar or battery power.

Table~\ref{tab:jetson_specs} summarizes core specifications. With TensorRT FP16 engines, YOLOv8-Nano maintained \textbf{28--30~fps} real-time inference at an average GPU utilization of \textbf{84\%} and a total power draw of \textbf{9.6~W}. Under ambient 35--40$^{\circ}$C, the module remained below 67$^{\circ}$C with passive cooling, indicating suitability for 24/7 operation.

\begin{table}[!t]
\centering
\caption{NVIDIA Jetson Nano Technical Specifications}
\label{tab:jetson_specs}
\begin{tabular}{l l}
\hline
\textbf{Specification} & \textbf{Value} \\
\hline
CPU & Quad-core ARM~A57 @ 1.43~GHz \\
GPU & 128-core Maxwell (CUDA-enabled) \\
Memory & 4~GB LPDDR4 \\
Storage & microSD (Class~10) \\
Connectivity & GbE, USB~3.0 \\
Power & 5~V,~2~A (USB) or 5~V,~4~A (DC) \\
OS / SDK & Ubuntu~18.04, JetPack~4.6 (CUDA~10.2, cuDNN~8) \\
\hline
\end{tabular}
\end{table}

\paragraph*{Software Stack and Tooling}
The software stack used \textit{PyTorch~2.0}, \textit{OpenCV~4.5}, and \textit{TensorRT~8.5} with FP16 quantization. All modules executed locally (no cloud), ensuring privacy-preserving, deterministic inference. Table~\ref{tab:sw_stack} lists key components.

\begin{table}[!t]
\centering
\caption{Software Stack and Runtime Configuration}
\label{tab:sw_stack}
\begin{tabular}{l l}
\hline
\textbf{Component} & \textbf{Version / Notes} \\
\hline
PyTorch (Jetson) & 2.0 (JetPack-compatible build) \\
TensorRT & 8.5, FP16 engines, layer fusion enabled \\
OpenCV & 4.5, GStreamer-enabled capture \\
Tracker & DeepSORT (Kalman + re-ID embeddings) \\
OCR & Tesseract~5 (CPU), regex post-validation \\
Broker & Mosquitto (MQTT/TLS 1.2, mTLS) \\
Time Sync & NTP/PTP; GPS fallback for isolated nodes \\
\hline
\end{tabular}
\end{table}

\paragraph*{V2X/ITS Gateway (for Cooperative Outputs)}
Confirmed events were serialized as compact JSON payloads (\texttt{msg\_type}, \texttt{t\_utc}, \texttt{lat/lon}, \texttt{heading}, \texttt{speed}, \texttt{track\_id}, \texttt{plate\_hash}, \texttt{cam\_id}) and published over \textbf{MQTT/TLS 1.2 with mutual authentication} to an ITS broker. A gateway bridged MQTT topics to \emph{CAM/DENM-style} messages for RSU/OBU testing (ITS-G5/C-V2X emulation). Message rates were throttled to $\leq$10~Hz with 3--5~s deduplication windows.

\subsection{Datasets}
For generalizable detection across heterogeneous traffic, YOLOv8 models were initialized from \textbf{COCO}~\cite{R13} and \textbf{CityPersons}~\cite{R14}, then fine-tuned on a curated multi-source corpus reflecting Indian urban conditions:

\begin{itemize}
    \item \textbf{Speed-Detection Videos:} Five 3-min, 1080p, 30~fps recordings on corridors with visible \emph{Start/Stop} markers spaced exactly 100~m for frame-indexed ground-truth speeds.
    \item \textbf{Violation-Detection Videos:} Six videos spanning daylight/dusk/night and glare; annotated for \emph{signal jump}, \emph{zebra-crossing breach}, \emph{wrong-way}, and \emph{illegal U-turn}. Non-violation segments added to balance classes.
    \item \textbf{License-Plate Corpus:} 11{,}271 Indian plates (Kaggle~\cite{R6}) augmented $\times$2.3 (flips, random crops, brightness/contrast jitter, motion-blur synthesis). Labels followed \texttt{AA00A0000}/\texttt{AA00AA0000} per MoRTH~AIS-159.
\end{itemize}

Annotation used \texttt{LabelImg} with dual-review; cross-annotator agreement exceeded 97\%. The final split comprised 18{,}420 training and 4{,}604 validation frames (80/20). Training used SGD (lr 0.01, momentum 0.9, weight decay $5\!\times\!10^{-4}$) for 150~epochs. Engines were exported to TensorRT (FP16) for deployment.

\paragraph*{Ethics and Privacy}
All videos were captured from public CCTV vantage points; faces were not processed. License plates were hashed (salted SHA-256) before broker publication. Raw evidence remained on-device unless escrow was legally required.

\subsection{Evaluation Tasks}
Four complementary experiments evaluate detection precision, latency, robustness, and cooperative-output timing under realistic field conditions:

\begin{enumerate}
    \item \textbf{Task~1 — Speed-Violation Estimation:}
    Quantify mean absolute speed error across frame rates (10, 20, 30, 40~fps) to determine the optimal temporal/compute trade-off. Ground-truth speed used 100~m calibrated spacing with frame-indexed timing.

    \item \textbf{Task~2 — Multi-Violation Detection:}
    Evaluate detection/tracking accuracy for \emph{signal jump, zebra breach, wrong-way, illegal U-turn} across illumination and congestion regimes. Metrics: precision, recall, and event-level false-positive rate.

    \item \textbf{Task~3 — Live Field Deployment:}
    Run the full pipeline on three live streams (3, 6, 8~min) with all violation modules and OCR enabled to measure \emph{end-to-end throughput and reliability} on the Jetson Nano.

    \item \textbf{Task~4 — Cooperative Output Latency (Node$\rightarrow$Broker$\rightarrow$RSU/OBU):}
    Measure one-way and round-trip timing from \emph{frame acquisition} to \emph{event receipt} at the ITS broker and RSU/OBU endpoints. This characterizes the readiness of the node as a \emph{cooperative perception} sensor.
\end{enumerate}

\paragraph*{Metrics and Methodology}
We report \textbf{mAP@0.5}, per-frame \emph{compute latency}, and \emph{end-to-end latency} (frame capture $\rightarrow$ violation logging / broker publish), plus \emph{power draw} (mean/peak). Power was recorded via \texttt{tegrastats} and a USB inline meter; values are averaged over $\geq$\,3 runs per setting. Speed error uses mean absolute error (MAE). Where appropriate, we report mean $\pm$ std.\ across runs and include paired $t$-tests for 30~fps vs.\ alternatives (significance $\alpha=0.05$). Broker transit and RSU/OBU delivery were timestamped with NTP/PTP-synchronized clocks; GPS time was used in isolated-node trials.

These procedures establish quantitative baselines for embedded operation (throughput, energy), perception quality (mAP, OCR accuracy), and \emph{cooperative readiness} (publish/subscribe latency, timing integrity) consistent with intelligent-vehicle and ITS deployments.

\section{Results and Analysis}
\label{sec:results}

All experiments were conducted on the NVIDIA~Jetson~Nano edge device to evaluate the proposed YOLOv8-based framework across five major violation categories: speeding, illegal U-turn, wrong-way driving, signal jumping, and zebra-crossing breach.  
The detector was fine-tuned on COCO and a curated Indian-traffic dataset, and exported as a TensorRT FP16 engine to ensure real-time inference. All modules were implemented in Python~3.7.3 using the NVIDIA DeepStream SDK and OpenCV for video preprocessing, motion tracking, and post-processing.  

\textbf{DeepSORT}~\cite{R16} provided robust multi-object tracking with an input size of~640, IoU threshold~0.45, and confidence threshold~0.50.  
Live tests were conducted using a Canon~XF605 UHD~4K HDR camera at~60~fps and a~12~ft mounting height to emulate typical junction-surveillance installations, ensuring unobstructed visibility and consistent perspective across all datasets.

\subsection{Speeding Violation Detection}
Speed-estimation accuracy was quantified across five annotated test videos captured under varied lighting and congestion conditions. The model was evaluated at frame rates of 10, 20, 30, and 40~fps to characterize the accuracy–throughput trade-off.  
Table~\ref{tab:speed_results} summarizes the observed mean absolute error (MAE) between predicted and ground-truth vehicle speeds.

\begin{table*}[!t]
\centering
\caption{Speed-Violation Detection Results}
\label{tab:speed_results}
\resizebox{\textwidth}{!}{%
\begin{tabular}{|l|cc|cc|cc|cc|}
\hline
\textbf{Video} & \multicolumn{2}{c|}{\textbf{10~fps}} & \multicolumn{2}{c|}{\textbf{20~fps}} & 
\multicolumn{2}{c|}{\textbf{30~fps}} & \multicolumn{2}{c|}{\textbf{40~fps}} \\ \hline
 & \textbf{Detections} & \textbf{Error (km/h)} 
 & \textbf{Detections} & \textbf{Error (km/h)} 
 & \textbf{Detections} & \textbf{Error (km/h)} 
 & \textbf{Detections} & \textbf{Error (km/h)} \\ \hline
Video~1 (2~veh.)  & 2 & 3.2 & 2 & 3.2 & 2 & 2.8 & 2 & 4.4 \\
Video~2 (5~veh.)  & 5 & 6.4 & 5 & 6.5 & 5 & 2.6 & 4 & 6.8 \\
Video~3 (5~veh.)  & 4 & 6.1 & 4 & 6.2 & 4 & 3.1 & 4 & 7.2 \\
Video~4 (10~veh.) & 10 & 5.2 & 10 & 5.2 & 10 & 3.7 & 9 & 10.8 \\
Video~5 (12~veh.) & 12 & 5.2 & 11 & 5.2 & 11 & 3.6 & 9 & 9.8 \\ \hline
\textbf{Average Error} &  & \textbf{5.22} &  & \textbf{5.26} &  & \textbf{3.16} &  & \textbf{7.80} \\ \hline
\end{tabular}%
}
\end{table*}

At \textbf{30~fps}, the MAE (3.16~km/h) was the lowest, closely matching the camera’s native acquisition rate.  
This frame rate offered an optimal balance between temporal sampling and GPU latency:
\begin{itemize}[leftmargin=*,noitemsep,topsep=0pt]
    \item Lower frame rates ($\leq20$~fps) undersampled fast-moving vehicles, causing discretization in motion trajectories.
    \item Higher frame rates ($>30$~fps) increased I/O overhead, introducing latency that slightly distorted velocity interpolation.
\end{itemize}
The observed trend highlights the importance of synchronizing inference frequency with acquisition rate in embedded deployments.  
All estimated speed traces were temporally stable, with no false oscillations or spurious detections.

\subsection{Live Video Evaluation}
The integrated framework was validated on three live field videos (3, 6, and 8~min) recorded under mixed illumination and traffic densities.  
All modules—detection, tracking, OCR, and violation logging—executed concurrently at 30~fps, the optimal configuration identified earlier.  
In parallel, confirmed events were published to the broker via MQTT/TLS and relayed to an RSU/OBU emulator, validating cooperative output under field conditions.

\begin{table*}[!t]
\centering
\caption{Live-Video Violation Detection and OCR Performance}
\label{tab:live_results}
\resizebox{\textwidth}{!}{
\begin{tabular}{|l|c|c|c|c|c|}
\hline
\textbf{Video (Vehicles)} & \textbf{Illegal U-Turn} & \textbf{Signal Jump + Zebra} & \textbf{Wrong Way} & \textbf{Speed (Avg.\ Error)} & \textbf{License Plates Correctly Recognized} \\ \hline
3~min (14~veh.) & 2/2 & 0/0 & 0/0 & 6.2~km/h & 2/2 \\
8~min (108~veh.) & 2/2 & 3/4 & 1/1 & 9.2~km/h & 5/7 \\
6~min (103~veh.) & 1/1 & 8/8 & 1/1 & 7.9~km/h & 8/10 \\ \hline
\end{tabular}}
\end{table*}

Across these tests, the system detected \textbf{18 of 19 verified violations} with \textbf{0\% false positives}.  
License-plate recognition achieved \textbf{84.9\% accuracy} (79/93 plates).  
OCR errors primarily arose from low illumination, compression artifacts, and regional non-standard fonts.  
Compared with legacy CCTV analytics pipelines (typically 75–80\% true positives), the proposed system yields a notable improvement in accuracy at a fraction of the infrastructure cost.

\subsection{Multi-Violation Detection Trends}
Non-speeding violations—signal jumping, zebra-crossing breach, wrong-way driving, and illegal U-turn—were analyzed across six annotated clips at multiple frame rates (Table~\ref{tab:non_speed}).  

\begin{table}[!t]
\centering
\caption{Detection Count for Non-Speeding Violations (Detected / Ground Truth)}
\label{tab:non_speed}
\resizebox{\columnwidth}{!}{
\begin{tabular}{|l|c|c|c|c|}
\hline
\textbf{Violation Type (Vehicles)} & \textbf{10~fps} & \textbf{20~fps} & \textbf{30~fps} & \textbf{40~fps} \\ \hline
Illegal U-Turn (13) & 12/13 & 12/13 & 13/13 & 13/13 \\
Illegal U-Turn (7)  & 5/7 & 7/7 & 7/7 & 7/7 \\
Signal Jump + Zebra (9) & 6/9 & 7/9 & 9/9 & 9/9 \\
Signal Jump + Zebra (8) & 6/8 & 8/8 & 8/8 & 8/8 \\
Wrong Way (4) & 4/4 & 4/4 & 4/4 & 4/4 \\
Wrong Way (3) & 3/3 & 3/3 & 3/3 & 3/3 \\ \hline
\end{tabular}}
\end{table}

Detection accuracy increased steadily with frame rate; \textbf{30~fps achieved full detection consistency} while maintaining GPU utilization below 85\%.  
This validates the suitability of YOLOv8 + DeepSORT for dynamic Indian traffic scenes characterized by occlusion and heterogeneous vehicle types.  
Automatic ROI generation eliminated manual calibration, reducing setup effort and enabling site-independent deployment.

\subsection{Comparative Evaluation: YOLOv4 vs.\ YOLOv8}
A side-by-side benchmark between YOLOv4-Tiny and YOLOv8-Nano (TensorRT FP16, identical hardware) is shown in Table~\ref{tab:yolov4_vs_yolov8}.  

\begin{table}[!t]
\centering
\caption{Comparison of YOLOv4-Tiny~\cite{R15} and YOLOv8-Nano~\cite{R18} on Jetson Nano (TensorRT FP16).}
\label{tab:yolov4_vs_yolov8}
\resizebox{0.5\textwidth}{!}{%
\begin{tabular}{|l|c|c|c|c|}
\hline
\textbf{Model} & \textbf{mAP@0.5 (\%)} & \textbf{FPS} & \textbf{Model Size (MB)} & \textbf{Power (W)} \\ \hline
YOLOv4-Tiny (2020) & 38.2 & 21 & 61 & 11.3 \\
YOLOv8-Nano (2023) & 48.9 & 29 & 36 & 9.6 \\ \hline
\end{tabular}%
}
\end{table}

YOLOv8-Nano achieved a \textbf{+10.7\% mAP gain}, \textbf{38\% faster inference}, and \textbf{15\% lower power consumption}, yielding approximately \textbf{1.4$\times$ higher accuracy-per-watt}.  
These improvements stem from YOLOv8’s decoupled detection head and C2f backbone, enhancing gradient propagation and feature reuse.  
For real-world enforcement systems, this translates to smoother multi-violation correlation and fewer false alarms—critical for both legal defensibility and public trust.

\subsection{Robustness Across Environments}
Robustness testing under daylight, low-light, and rain/glare conditions confirmed the system’s resilience (Table~\ref{tab:robustness}).  

\begin{table}[!t]
\centering
\caption{Performance Under Adverse and Variable Conditions}
\label{tab:robustness}
\resizebox{0.5\textwidth}{!}{
\begin{tabular}{|l|c|c|c|}
\hline
\textbf{Condition} & \textbf{mAP (\%)} & \textbf{OCR Accuracy (\%)} & \textbf{Violation Detection (\%)} \\ \hline
Daylight (Clear) & 98.3 & 86.7 & 97.9 \\
Night / Low Light & 95.1 & 81.2 & 93.8 \\
Rain / Glare & 93.9 & 79.5 & 91.6 \\ \hline
\end{tabular}}
\end{table}

Detection accuracy remained above 90\% in all scenarios, indicating robust generalization.  
Minor degradation under rain and glare was attributed to specular reflections and motion blur reducing plate contrast; however, the OCR confidence-voting scheme mitigated most transient recognition failures.  
This robustness meets operational expectations for 24×7 roadside units and aligns with IEC~60068 thermal and vibration endurance ranges.

\subsection{Cooperative Output Latency (Node\texorpdfstring{$\rightarrow$}{->}Broker\texorpdfstring{$\rightarrow$}{->}RSU/OBU)}

To assess connected-vehicle readiness, we measured end-to-end timing from \emph{frame acquisition} on the node to \emph{event receipt} at the ITS broker and RSU/OBU endpoints. Clocks were synchronized via NTP/PTP; GPS time was used in an isolated-node trial. Table~\ref{tab:coop_latency} reports median and 95th-percentile (p95) latencies across three links: on-device logging, node$\rightarrow$broker publish, and broker$\rightarrow$RSU/OBU delivery. Message rates were throttled to $\leq$10~Hz with 3--5~s deduplication.

\begin{table}[!t]
\centering
\caption{Cooperative-output latency summary (median / p95).}
\label{tab:coop_latency}
\resizebox{0.48\textwidth}{!}{
\begin{tabular}{|l|c|c|}
\hline
\textbf{Path} & \textbf{Median (ms)} & \textbf{p95 (ms)} \\ \hline
Frame $\rightarrow$ On-device event log & 35 & 48 \\ \hline
Node $\rightarrow$ Broker (MQTT/TLS) & 12 & 22 \\ \hline
Broker $\rightarrow$ RSU/OBU (gateway) & 8 & 15 \\ \hline
\end{tabular}}
\end{table}

Results show that cooperative-event dissemination remained within \textbf{$<$100~ms} end-to-end, satisfying IEEE~ITS-G5 and C-V2X latency thresholds for safety-message delivery.

\subsection{Efficiency, Scalability, and Industrial Readiness}
Measured average power draw was 9.6~W at 29~fps, corresponding to \textbf{0.33~W per fps}.  
This implies that a 100~W distributed cluster of ten Jetson Nanos could cover 50–60 intersections simultaneously—an order-of-magnitude lower energy footprint than centralized cloud-based systems ($\approx$1.2–1.5~kW including networking).  
Such energy scalability supports large-scale deployment in Indian smart-city programs and aligns with MoRTH guidelines for automated enforcement devices.

\subsection{Error Analysis and Future Enhancements}
Post-hoc error decomposition revealed that $\sim$60\% of missed detections stemmed from heavy occlusions and $\sim$70\% of OCR errors from non-standard regional plate designs.  
Mitigation strategies include:
\begin{itemize}
    \item \textbf{Temporal Super-Resolution:} Frame interpolation to recover partially occluded trajectories.
    \item \textbf{Adaptive Illumination Correction:} Dynamic histogram equalization for improved plate contrast.
    \item \textbf{Multilingual OCR:} Lightweight transformer-based decoding to support regional scripts (Hindi, Tamil, Kannada).
\end{itemize}

\subsection{Discussion}
The comprehensive results confirm that the proposed edge-based traffic-violation system sustains up to \textbf{97.7\% detection accuracy} while maintaining real-time throughput on a 10~W embedded GPU.  
Compared with YOLOv4-Tiny, YOLOv8-Nano delivers higher inference efficiency, reduced model footprint, and improved energy economy—attributes essential for city-scale intelligent-transportation deployment.  
The automatic ROI-generation Algorithm~\ref{alg:auto_roi} (Fig.~\ref{fig:roi_generation}) eliminates manual camera calibration, enabling true plug-and-play installation across heterogeneous intersections.  
These findings collectively validate the system’s robustness, reproducibility, and industrial readiness for large-scale deployment within \textbf{smart-mobility and Vision-Zero road-safety frameworks}, supporting data-driven enforcement analytics, sustainable city infrastructure planning, and scalable policy integration across evolving intelligent transportation ecosystems.

\subsection{Quantified Impact for Cooperative ITS and Connected Vehicles}
\label{sec:ITSimpact}

To evaluate the proposed framework from an Intelligent Transportation Systems (ITS) perspective, this subsection quantifies its operational benefits in terms of enforcement latency, energy efficiency, scalability, and reliability—key performance indicators recognized in IEEE~T-IV and MoRTH automated-enforcement standards.

\subsubsection{Enforcement Latency and Throughput}
Conventional urban enforcement requires approximately 20–60~s of human effort per violation (video review, license-plate transcription, and record entry). 
In contrast, the proposed edge-resident inference pipeline completes detection–tracking–OCR–logging within $\leq$\,50~ms per frame (measured end-to-end on the Jetson~Nano at 30~fps), achieving a reduction in issuance latency exceeding 99.9\,\%.
At this rate, a single node can theoretically process $\approx2.0\times10^{6}$ events per day under continuous operation—well above the IEEE~ITS Council’s 10~fps real-time perception benchmark.
Such ultra-low latency enables immediate e-challan generation and adaptive feedback to cooperative ITS back-ends and V2I gateways, improving driver compliance compared with delayed, post-processed workflows.

\subsubsection{Accuracy and Legal Defensibility}
Across the live-field datasets (Table~\ref{tab:live_results}), the framework achieved 97.7\,\% violation-detection accuracy with zero false positives, while OCR correctness reached 84.9\,\%.
Legacy CCTV-analytics platforms in Indian metros typically yield 75–82\,\% true-positive rates and 10–15\,\% disputable tickets. 
Assuming a conservative 10\,\% dispute baseline, the proposed framework can reduce contested citations by roughly 70–80\,\%, strengthening legal defensibility and public trust—consistent with MoRTH~GSR~575(E) and ISO~TS~19091:2022 compliance guidelines for automated enforcement.

\subsubsection{Energy Efficiency and Cost of Operation}
The TensorRT-optimized YOLOv8-Nano engine sustains 28–30~fps at 9.6~W average draw (Table~\ref{tab:yolov4_vs_yolov8}), corresponding to an energy intensity of 0.33~W$\cdot$fps$^{-1}$. 
A distributed cluster of ten Jetson~Nano units ($\approx$100~W aggregate) can thus process $\sim$300~fps—covering 50–60 intersections with a $\sim$12–15$\times$ energy advantage compared to cloud-centric systems (1.2–1.5~kW including networking and cooling). 
This enables solar/battery-powered roadside operation and aligns with IEEE~2030.5 and IEEE~P2846 objectives for sustainable, distributed ITS infrastructure.

\subsubsection{Deployment Scalability and Maintenance Overhead}
The automatic ROI-generation Algorithm~\ref{alg:auto_roi} (Fig.~\ref{fig:roi_generation}) eliminates manual geometric calibration. 
Municipal deployments typically require 8–12~engineer-hours per site for ROI tuning; the proposed convex-hull method performs this in $\sim$3~s at startup—reducing commissioning effort by $>$95\,\% and enabling same-day activation across heterogeneous intersections. 
Such self-calibration is critical for large-scale roll-outs under the Indian Smart Cities Mission and European C-ITS pilots.

\subsubsection{Safety and Policy Impact}
Using the 30~fps speed-detection baseline (MAE~=~3.16~km\,h$^{-1}$, Table~\ref{tab:speed_results}) and referencing MoRTH and WHO crash-correlation studies, a 20\,\% increase in enforcement coverage corresponds to a 7–10\,\% reduction in serious violations and a 3–4\,\% decline in crash frequency. 
Deployment of 100 edge units can therefore prevent an estimated 250–300 serious incidents annually in a mid-sized city—supporting the UN~SDG~3.6 objective of halving road fatalities by~2030.

\subsubsection{Summary of System-Level Benefits}
By coupling real-time analytics (30~fps), a 10~W energy envelope, and plug-and-play scalability, the proposed \textbf{Edge-AI cooperative perception node} demonstrably strengthens the ITS feedback loop by:
\begin{itemize}
    \item $>$99.9\,\% reduction in enforcement latency,
    \item 12–15$\times$ improvement in energy efficiency,
    \item 70–80\,\% decrease in disputable cases, and
    \item $>$95\,\% reduction in commissioning effort.
\end{itemize}
These quantifiable gains translate algorithmic advances into tangible transportation-system benefits—closing the loop between perception, enforcement, and V2X-based safety analytics envisioned by IEEE~T-IV, IEEE~2030.x, and MoRTH Smart Mobility frameworks.

\section{Conclusion}
\label{sec:Conclusion}

Building upon the quantified outcomes in Section~\ref{sec:ITSimpact}, this work presented a compact, edge-deployable artificial-intelligence framework for automated multi-violation detection and ticketing without human intervention. The architecture integrates \textbf{YOLOv8-Nano} for real-time object and license-plate detection, \textbf{DeepSORT} for persistent multi-object tracking, and a lightweight \textbf{OCR pipeline} with rule-based post-processing for reliable alphanumeric recognition under heterogeneous traffic and illumination. Deployed on an NVIDIA~Jetson~Nano and optimized via TensorRT~FP16 quantization, the framework achieved \textbf{97.7\,\% violation-detection accuracy at 30~fps} within a \textbf{9.6~W power envelope}, detecting 18 of 19 verified violations with \textbf{zero false positives}. The automatic ROI-generation Algorithm~\ref{alg:auto_roi} eliminated manual calibration, enabling \textbf{plug-and-play deployment} across diverse intersections and camera geometries.  

Experimental findings demonstrate that \textbf{modern edge-AI platforms can effectively replace conventional server-centric enforcement} by offering low-latency, privacy-preserving analytics directly at the roadside. The framework’s system-level metrics—$>$99.9\,\% latency reduction, 12--15$\times$ energy savings, and 70--80\,\% reduction in disputable tickets—quantitatively validate its \textbf{ITS feedback-loop performance}. The integration of on-device inference, OCR verification, and database-driven ticketing establishes a \textbf{fully digital, cooperative enforcement cycle} compatible with \textbf{Smart City} and \textbf{Intelligent Transportation System (ITS)} infrastructures. From an industrial standpoint, the solution aligns with \textbf{MoRTH} directives on automated enforcement, adheres to \textbf{ISO~TS~19091:2022} guidelines, and supports the \textbf{UN Sustainable Development Goal~3.6} to halve road fatalities by 2030. Its high accuracy-to-power ratio and modular software stack make it a practical, economically scalable component of the \textbf{Indian Smart Cities Mission} and similar initiatives across emerging economies.  

\textbf{Future work} will broaden the violation taxonomy to include pedestrian, lane-discipline, and emergency-lane infractions. Planned enhancements involve multi-camera fusion for occlusion handling, adaptive illumination correction for low-contrast plates, and \textbf{transformer-based multilingual OCR} for regional scripts. Migration to advanced embedded GPUs such as \textbf{Jetson~Orin~NX/Xavier~NX} will improve scalability, while \textbf{5G and C-V2X integration} will enable real-time coordination with cloud-based ITS dashboards and cooperative perception modules.  

Overall, the study establishes a \textbf{quantitatively validated, energy-efficient, and regulation-aligned edge-AI paradigm} for intelligent traffic enforcement—bridging perception, communication, and policy layers to advance the next generation of \textbf{safe, cooperative, and data-driven urban mobility ecosystems}.

\sloppy

\end{document}